\documentclass[11pt,a4paper]{article}
\usepackage[hyperref]{eacl2021}
\usepackage{times}
\usepackage{latexsym}
\usepackage{adjustbox}
\usepackage{xspace}
\usepackage{caption}
\usepackage{amsmath}
\usepackage{amsfonts}
\usepackage{booktabs}

\usepackage{microtype}

\usepackage{cleveref}
\crefname{section}{\S}{\S\S}
\Crefname{section}{\S}{\S\S}
\crefname{table}{Tab.}{}
\crefname{figure}{Fig.}{}
\crefname{algorithm}{Alg.}{}
\crefname{equation}{eq.}{}
\crefname{appendix}{App.}{}
\crefformat{section}{\S#2#1#3}  %

\usepackage{todonotes}

\aclfinalcopy %

\newcommand{\insertOptimTable}{
\begin{table}[t]
\centering
\begin{adjustbox}{width=0.8\columnwidth}
\begin{tabular}{ll| cc}
\toprule
LS & $\beta_2$ & Vanilla & Feature Invariant\\
\midrule
0 & 0.999 & 89.34 & 89.80 \\
0 & 0.98 & 89.62 & 89.92 \\
0.1 & 0.999 & 89.48 & 90.02 \\
0.1 & 0.98 & 89.98 & 90.28 \\
\bottomrule
\end{tabular}
\end{adjustbox}
\caption{Average development accuracy on morphological inflection with different LS and $\beta_2$, which denote hyperparameter of label smoothing and Adam optimizer respectively. }
\label{table:optim}
\end{table}}

\newcommand{\insertCONLLTable}{
\begin{table}[t]
\centering
\begin{adjustbox}{width=\columnwidth}
\begin{tabular}{l ll}
\toprule
 & ACC & Dist \\
\midrule
\newcite{silfverberg-etal-2017-data}* & 92.97 & 0.170 \\
\newcite{wu-etal-2018-hard} & 93.60 & 0.128 \\
\newcite{wu-cotterell-2019-exact} & 94.40 & 0.113 \\
\newcite{wu-cotterell-2019-exact} (Our eval) & 94.81 & 0.123 \\
\newcite{makarov-etal-2017-align}* & 95.12 & 0.100 \\
\newcite{bergmanis-etal-2017-training}* & 95.32 & 0.100 \\
\midrule
Transformer (Dropout = 0.3) & \textbf{95.59} & \textbf{0.088} \\
Transformer (Dropout = 0.1) & 95.56 & 0.090 \\
\bottomrule
\end{tabular}
\end{adjustbox}
\caption{Average test performance on morphological inflection of Transformer against models from the literature. ${}^*$ denotes model ensembling.}
\label{table:morph-inflect}
\end{table}}

\newcommand{\insertHistTable}{
\begin{table}[t]
\centering
\begin{adjustbox}{width=1\columnwidth}
\begin{tabular}{l llll}
\toprule
 & ACC & CER$_i$ & ACC$^s$ & CER$^s_i$ \\
\midrule
\newcite{ljubesic16-normalising} & 91.78 & 0.392 & 90.37 & 0.360 \\
\newcite{ljubesic16-normalising} (LM) & 91.56 & 0.399 & 89.93 & 0.368 \\
\newcite{bollmann2018normalization} & 91.27 & 0.381 & 89.73 & 0.350 \\
\newcite{tang-etal-2018-evaluation} & 91.67 & 0.389 & 90.32 & 0.358 \\
\newcite{flachs-etal-2019-historical} & - & - & 90.06 & - \\
\midrule
Transformer (Dropout = 0.3) & 91.30 & \textbf{0.340} & 89.99 & \textbf{0.330} \\
Transformer (Dropout = 0.1) & \textbf{91.85} & 0.352 & \textbf{90.61} & 0.334 \\
\bottomrule
\end{tabular}
\end{adjustbox}
\caption{Average test performance on historical text normalization of Transformer against models from the literature. $^s$ denote subset of dataset as \newcite{flachs-etal-2019-historical} only experiment with subset of languages.}
\label{table:hist-norm}
\end{table}}

\newcommand{\insertOtherTable}{
\begin{table}[t]
\centering
\begin{adjustbox}{width=1\columnwidth}
\begin{tabular}{l llll}
\toprule
 & WER & PER & ACC & MFS\\
\midrule
\newcite{wu-etal-2018-hard} & 28.20 & \textbf{0.068} & 41.10 & 0.894\\
\newcite{wu-cotterell-2019-exact} & 28.20 & 0.069 & 41.20 & 0.895\\
\midrule
Transformer (Dropout = 0.3) & 28.08 & 0.070 & \textbf{43.39} & \textbf{0.897} \\
Transformer (Dropout = 0.1) & \textbf{27.63} & 0.069 & 41.35 & 0.891 \\
\bottomrule
\end{tabular}
\end{adjustbox}
\caption{Average test performance on Grapheme-to-Phoneme and dev performance on Transliteration of Transformer against models from the literature.}
\label{table:g2p-trans}
\end{table}}

\title{Applying the Transformer to Character-level Transduction}

\usepackage{tipa}

\newcommand{\jhu}{\normalfont \text{\textipa{Z}}}
\newcommand{\ethz}{\text{\normalfont \textipa{Q}}}
\newcommand{\ucambridge}{\normalfont \text{\textipa{6}}}
\newcommand{\cub}{\textrm{\normalfont \textipa{X}}}
\newcommand\blfootnote[1]{%
  \begingroup
  \renewcommand\thefootnote{}\footnote{#1}%
  \addtocounter{footnote}{-1}%
  \endgroup
}
  
\author{Shijie Wu$^{\jhu}$~~~\;~~~\textbf{Ryan Cotterell}$^{\ethz,\ucambridge}$~~~\;~~~\textbf{Mans Hulden}$^{\cub}$ \\
$^{\jhu}$Johns Hopkins University~~~\;~~~$^{\ucambridge}$University of Cambridge\\
$^{\ethz}$ETH Z{\"u}rich~~~\;~~~$^\cub$University of Colorado Boulder\\
\texttt{shijie.wu@jhu.edu}~\;~\texttt{ryan.cotterell@inf.ethz.ch}~\;~\texttt{mans.hulden@colorado.edu}}

\date{}

\begin{document}
\maketitle
\begin{abstract}
The transformer \cite{vaswani2017attention} has been shown to outperform recurrent neural network-based sequence-to-sequence models in various word-level NLP tasks.
Yet for character-level transduction tasks, e.g. morphological inflection generation and historical text normalization, there are few works that outperform recurrent models using the transformer.
In an empirical study, we uncover that, in contrast to recurrent sequence-to-sequence models, the batch size plays a crucial role in the performance of the transformer on character-level tasks, and we show that with a large enough batch size, the transformer does indeed outperform recurrent models. We also introduce a simple technique to handle feature-guided character-level transduction that further improves performance.
With these insights, we achieve state-of-the-art performance on morphological inflection and historical text normalization. We also show that the transformer outperforms a strong baseline on two other character-level transduction tasks: grapheme-to-phoneme conversion and transliteration.
\blfootnote{Code will be available at \url{https://github.com/shijie-wu/neural-transducer}.}
\end{abstract}

\section{Introduction}
The transformer \cite{vaswani2017attention} has become a popular architecture for sequence-to-sequence transduction in NLP. It has achieved state-of-the-art performance on a range of common word-level transduction tasks: neural machine translation \cite{barrault-etal-2019-findings}, question answering \cite{devlin-etal-2019-bert} and abstractive summarization \cite{dong2019unified}. In addition, the
transformer forms the backbone of the widely-used BERT \cite{devlin-etal-2019-bert}. 
Yet for character-level transduction tasks like morphological inflection, the dominant model has remained a recurrent neural network-based sequence-to-sequence model with attention \cite{cotterell-etal-2018-conll}. This is not for lack of effort---but rather, it is the case that the transformer has consistently underperformed in experiments on average \cite{tang-etal-2018-self}.\footnote{This claim is also based on the authors' personal communication with other researchers in morphology in the corridors of conferences and through email.}
As anecdotal evidence of this, we note that in the 2019 SIGMORPHON shared task on cross-lingual transfer for morphological inflection, no participating system was based on the transformer \cite{mccarthy-etal-2019-sigmorphon}.%

\begin{figure}
\centering
\includegraphics[width=1\columnwidth]{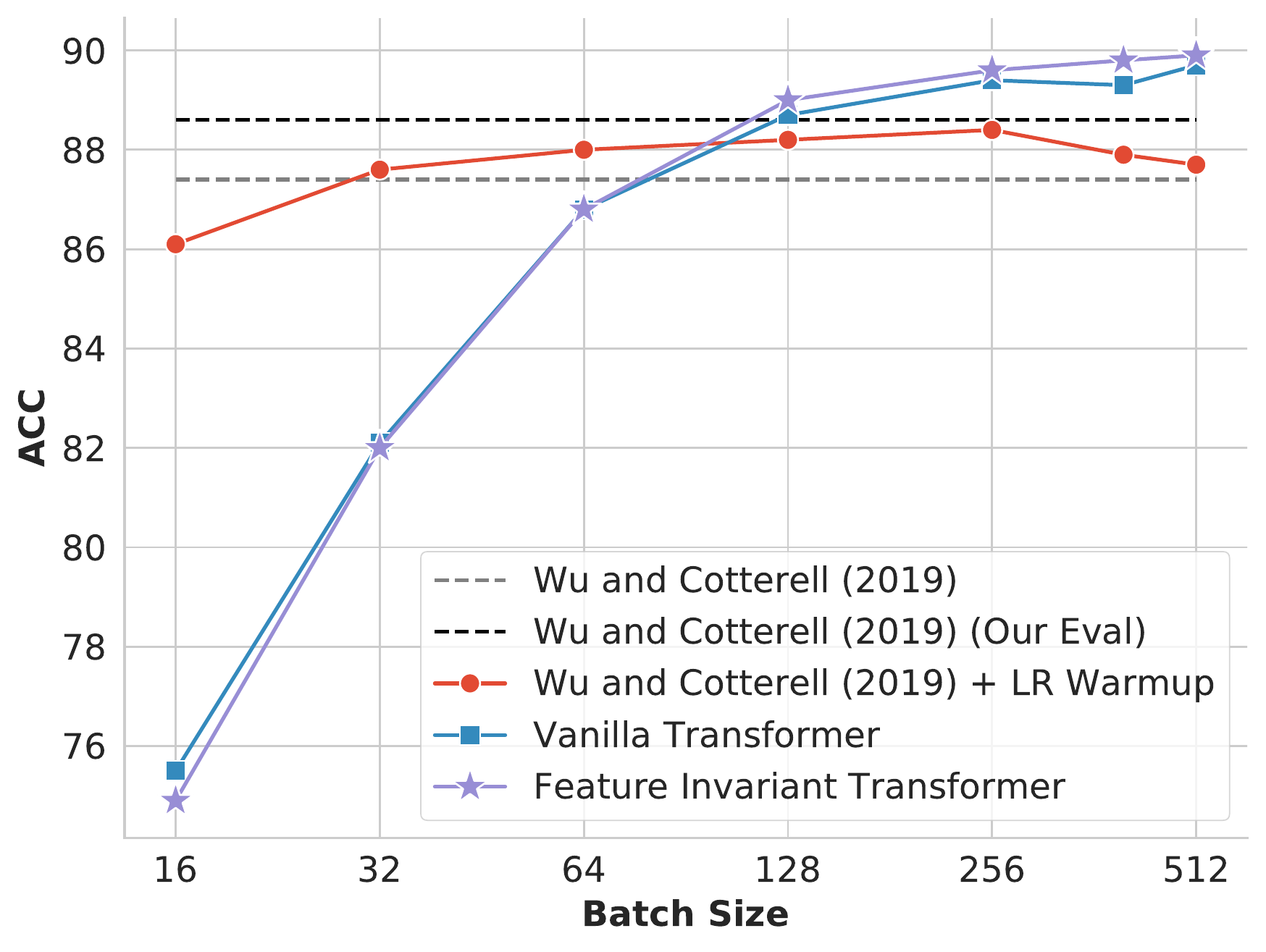}
\caption{Development set accuracy for 5 languages on morphological inflection with different batch sizes. We evince our two primary contributions: (1) we set the \textbf{new state of the art} morphological inflection using the transformer and (2) we demonstrate the transformer's \textbf{dependence on the batch size}.}
\label{fig:batch-size}
\end{figure}

\begin{figure*}
\centering
\includegraphics[width=2\columnwidth]{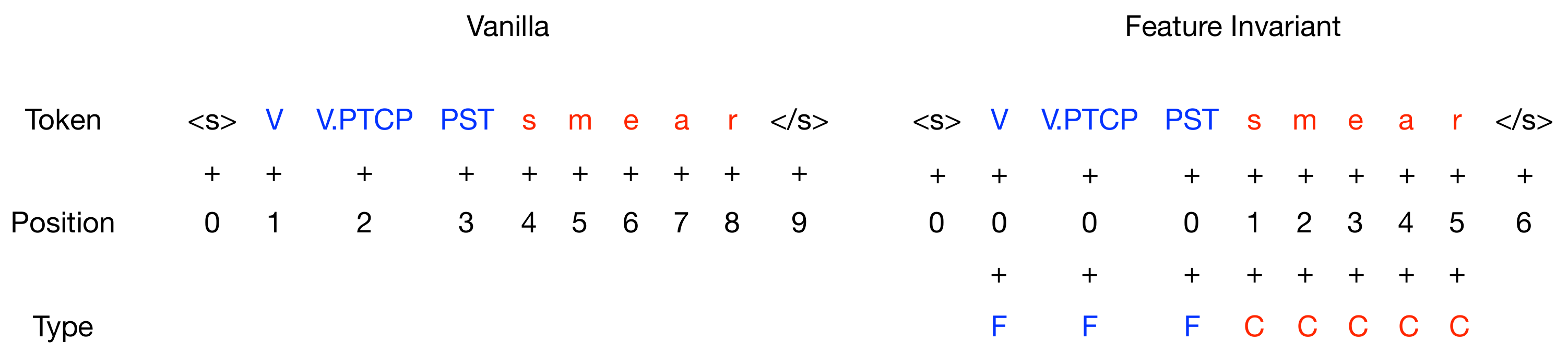}
\caption{Handling of feature-guided character-level transduction with special position and type embeddings in the encoder. \texttt{F} denotes features while \texttt{C} denotes characters. We use morphological inflection as an example, inflecting \textit{smear} into its past participle form, \textit{smeared}.}
\label{fig:position}
\end{figure*}

Character-level transduction models are often trained with less data than their word-level counterparts:
In contrast to machine translation, where millions of training samples are available, the 2018 SIGMORPHON shared task \cite{cotterell-etal-2018-conll} high-resource setting only provides $\approx$ 10k training examples per language.
It is also not obvious that non-recurrent architectures such as the transformer should provide an advantage at many character-level tasks:
For instance, \newcite{gehring2017} and \newcite{vaswani2017attention} suggest that transformers (and convolutional models in general) should be better at remembering long-range dependencies.
In the case of morphology, none of these considerations seem relevant: inflecting a word (a) requires little capacity to model long-distance dependencies and is largely a monotonic transduction; (b) it involves no semantic disambiguation, the tokens in question being letters; (c) it is not a task for which parallelization during training appears to help, since training time has never been an issue in morphology tasks.\footnote{Many successful CoNLL--SIGMORPHON shared task participants report training their models on laptop CPUs.}%

In this work, we provide state-of-the-art numbers for
morphological inflection and historical text normalization, 
a novel result in the literature. We also show the transformer
outperforms a strong recurrent baseline on two other character-level tasks: grapheme-to-phoneme (g2p) conversion
and transliteration. We find 
that a single hyperparameter, batch size, is largely responsible for the previous poor results. Despite having fewer parameters, the transformer outperforms the recurrent sequence-to-sequence baselines on all four tasks. We conduct
a short error analysis on the task of morphological inflection
to round out the paper.

\section{The Transformer for Characters}

\paragraph{The Transformer.}
The transformer, originally described by \newcite{vaswani2017attention}, is a self-attention-based encoder-decoder model. The encoder has $N$ layers, consisting of a multi-head self-attention layer and a two-layer feed-forward layer with ReLU activation, both equipped with a skip connection. The decoder has a similar structure as the encoder except that, in each decoder layer between the self-attention layer and feed-forward layer, a multi-head attention layer attends to the output of the encoder. Layer normalization \cite{ba2016layer} is applied to the output of each skip connection. Sinusoidal positional embeddings are used to incorporate positional information without the need for recurrence or convolution. Here, we describe two modifications
we make to the transformer for character-level tasks.%

\paragraph{A Smaller Transformer.} 
As the dataset sizes in character-level transduction tasks are significantly smaller than in machine translation, we employ a smaller transformer with $N=4$ encoder-decoder layers. We use 4 self-attention heads. The embedding size is $d_{\textit{model}} = 256$ and the hidden size of the feed-forward layer is $d_{\textit{FF}} = 1024$. In the preliminary experiments, we found that using layer normalization before self-attention and the feed-forward layer performed slightly better than the original model. It is also the default setting of a popular implementation of the transformer \cite{vaswani2018tensor2tensor}. The transformer alone has around 7.37M parameters, excluding character embeddings and the linear mapping before the softmax layer. We decode the model left to right in a greedy fashion. 

\paragraph{Feature Invariance.} 
Some character-level transduction is guided by features. 
For example, in the case of morphological reinflection, the task requires a set of morphological attributes that control what form a citation form is inflected into (see \cref{fig:position} for an example). 
However, the order of the features is irrelevant. In a recurrent neural network, features are input in some predefined order as special characters and pre- or postpended to the input character sequence representing the citation form. The same is true for a vanilla transformer model, as shown on the left-hand side of \cref{fig:position}. This leads to different relative distances between a character and a set of features.\footnote{While the features could be encoded with a binary vector followed by MLP, it introduces a representation bottleneck for encoding features.} To avoid such an inconsistency, we propose a simple remedy: We set the positional encoding of features to 0 and only start counting the positions for characters. Additionally, we add a special token to indicate whether a symbol is a word character or a feature. The right-hand side of \cref{fig:position} evinces how we have the same relative distance between characters and features.%

\section{Empirical Findings}

\paragraph{Tasks.} 
We consider four character-level transduction tasks: morphological inflection, grapheme-to-phoneme conversion, transliteration, and historical text normalization. For morphological inflection, we use the 2017 SIGMORPHON shared task data \cite{cotterell-etal-2017-conll} with 52 languages. The performance is evaluated by accuracy (ACC) and edit distance (Dist). For the g2p task, we use the unstressed CMUDict \cite{CMUDict} and NETtalk \cite{Sejnowski1987ParallelNT} resources. We use the splits from \newcite{wu-etal-2018-hard}. We evaluate under word error rate (WER) and phoneme error rate (PER). For transliteration, we use the NEWS 2015 shared task data \cite{zhang-etal-2015-whitepaper}.\footnote{We do not have access to the test set.} For historical text normalization, we follow \newcite{bollmann-2019-large} and use datasets for Spanish \cite{sanchez2013open}, Icelandic and Swedish \cite{pettersson2013smt}, Slovene \cite{scherrer2013modernizing,scherrer2016modernising,ljubevsic2016normalising}, Hungarian and German \cite{pettersson2016spelling}.\footnote{We do not include English due to licensing issues.} We evaluate using accuracy (ACC) and character error rate of incorrect prediction (CER$_i$).
\begin{figure}
\centering
\includegraphics[width=1\columnwidth]{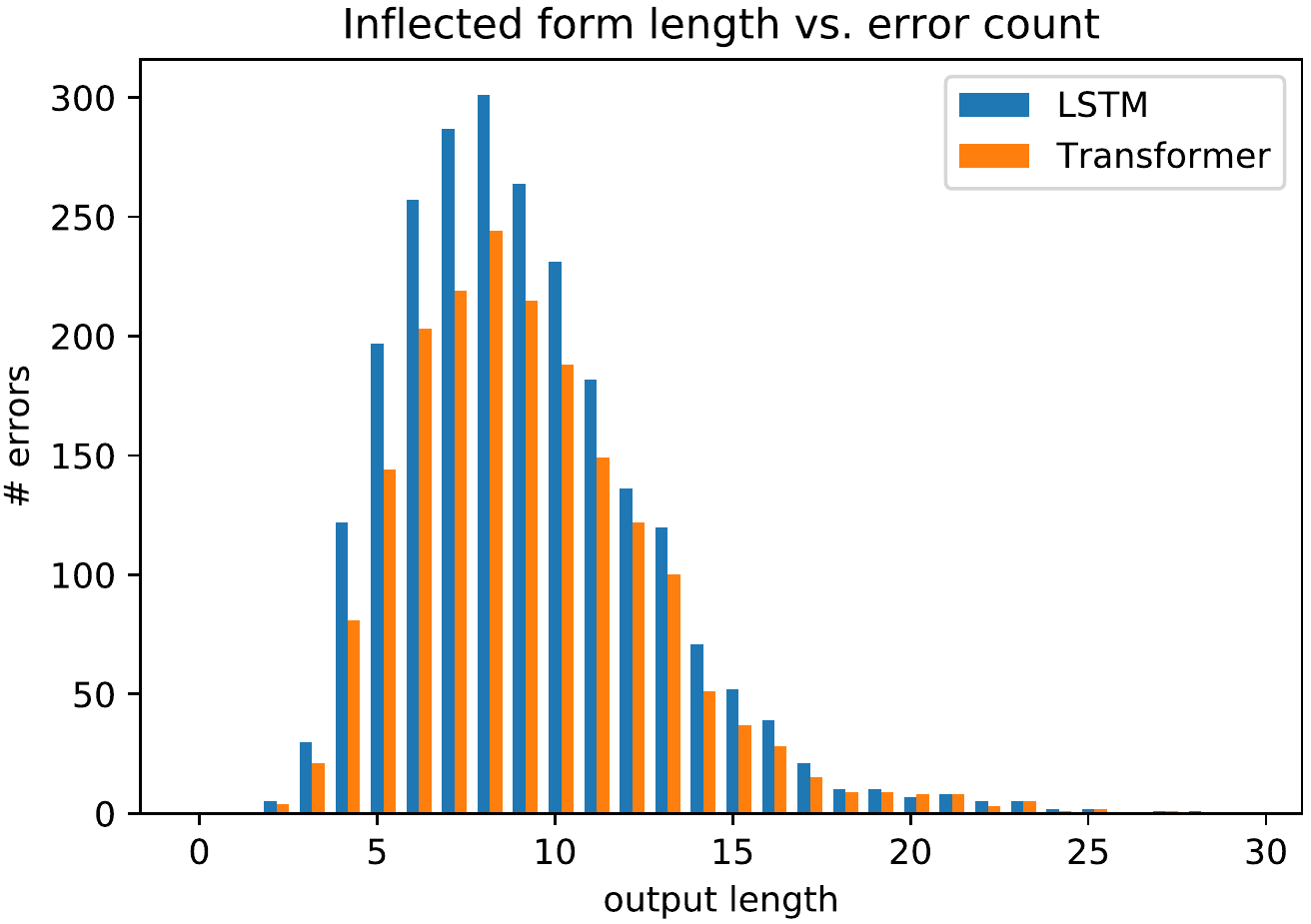}
\caption{Distribution of incorrectly inflected forms in the test set of the inflection task over all 52 languages grouped by desired output word length.}
\label{fig:errorlengths}
\end{figure}
\insertOptimTable

\paragraph{Optimization.} 
We use Adam \cite{kingma2014adam} with a learning rate of $0.001$ and an inverse square root learning rate scheduler \cite{vaswani2017attention} with 4k steps during the warm-up. We train the model for 20k gradient updates and save and evaluate the model every 400 gradient updates. We select the best model out of 50 checkpoints based on development set accuracy. The number of gradient updates and checkpoints are roughly the same as \newcite{wu-cotterell-2019-exact}, the single model state of the art on the 2017 SIGMORPHON dataset. We use their model as a baseline model. For all experiments, we use a single predefined random seed.

\subsection{A Controlled Hyperparameter Study}
\label{sec:ablation}
To demonstrate the importance of hyperparameter
tuning for the transformer on character-level tasks, we perform a small controlled hyperparameter study. This is important since researchers had previously failed to achieve high-performing results with the transformer on
character-level tasks. Here, we look at morphological inflection on the five languages in the 2017 SIGMORPHON dataset where submitted systems performed the worst: Latin, Faroese, French, Hungarian, and Norwegian (Nynorsk). We set the dropout to 0.3, $\beta_2$ of Adam to 0.999 (the default value), and do not use label smoothing. We do not tune any other hyperparameter except the following three hyperparameters.

\paragraph{The Importance of Batch Size.}
While recurrent models like \citeauthor{wu-cotterell-2019-exact} use a batch size of 20, halving the learning rate when stuck and employing early stopping, we find that a less aggressive learning rate scheduler, allowing the model to train longer, outperforms these hyperparameters. \cref{fig:batch-size} shows that the \emph{significant impact of batch size on the transformer}. The transformer performance increases steadily as the batch size is increased, similarly to what \newcite{popel2018training} observe for machine translation. The transformer only outperforms the recurrent baseline when the batch size is at least 128, which is much larger than batch size commonly used in recurrent models.\footnote{It is also large in the context of character-level tasks, which typically have around 10k training examples. Batch size of 400 would imply approximately 4\% of training data in a single gradient update.} Note that the model of \citeauthor{wu-cotterell-2019-exact} has 8.66M parameters, 17\% more than the transformer model. To get an apples-to-apples comparison, we apply the same learning rate scheduler to \citeauthor{wu-cotterell-2019-exact}; this does not yield similar improvements and underperforms with respect to the traditional learning rate scheduler. Our feature invariant transformer also outperforms the vanilla transformer model. We set the batch size to 400 for our main experiments. Note the batch size of 400 is especially large (4\% of training data) considering the training size is only 10k.

\paragraph{Other Hyperparameters.} \newcite{vaswani2017attention} applies label smoothing \cite{szegedy2016rethinking} of 0.1 to the transformer model and shows that it hurts perplexity, but improves BLEU scores for machine translation. Instead of the default 0.999 $\beta_2$ for Adam, \newcite{vaswani2017attention} uses 0.98 and we find that both choices benefit character-level transduction tasks as well (see \cref{table:optim}).

\insertCONLLTable

\insertHistTable

\insertOtherTable

\subsection{New State-of-the-Art Results}
We train our feature invariant transformer on the four character-level tasks, exhibiting state-of-the-art results on morphological inflection and historical text normalization.

\paragraph{Morphological Inflection.} 
As shown in \cref{table:morph-inflect}, the feature invariant transformer produces state-of-the-art results on the 2017 SIGMORPHON shared tasks, improving upon ensemble-based systems by 0.27 points. We observe that as the dataset decreases in size, a model with a larger dropout value performs slightly better. A brief tally of phenomena that are difficult to learn for many machine learning models, categorized along typical linguistic dimensions (such as word-internal sound changes, vowel harmony, circumfixation, ablaut, and umlaut phenomena) fail to reveal any consistent pattern of advantage to the transformer model. In fact, errors seem to be randomly distributed with an overall advantage of the transformer model. Curiously, errors grouped along the dimension of word length reveal that as word forms grow longer, the transformer advantage shrinks (\cref{fig:errorlengths}).

\paragraph{Historical Text Normalization.} 
\cref{table:hist-norm} shows that the transformer model with dropout of 0.1, as in the case of morphological inflection, improves upon the previous state of the art, although the model with a dropout of 0.3 yields a slightly better CER$_i$.%

\paragraph{G2P and Transliteration.} 
\cref{table:g2p-trans} shows that the transformer outperforms previously published strong recurrent models on two tasks despite having fewer parameters. A dropout rate of 0.3 yields significantly better performance on the transliteration task while a dropout rate of 0.1 is stronger on the g2p task. This shows that transformers can and do outperform recurrent transducers on common character-level tasks when properly tuned.

\section{Related Work}
Character-level transduction is largely dominated by attention-based LSTM sequence-to-sequence \cite{luong-etal-2015-effective} models \cite{cotterell-etal-2018-conll}.
Character-level transduction tasks usually involve input-output pairs that share large substrings and alignments between these are often monotonic. Models that address the task tend to focus on exploiting such structural bias.
Instead of learning the alignments, \newcite{aharoni-goldberg-2017-morphological} use external monotonic alignments from the SIGMORPHON 2016 shared task baseline \newcite{cotterell-etal-2016-sigmorphon}. \newcite{makarov-etal-2017-align} use this approach to win the CoNLL-SIGMORPHON 2017 shared task on morphological inflection \citep{cotterell-etal-2017-conll}.
\newcite{wu-etal-2018-hard} shows that explicitly modeling alignment (hard attention) between source and target characters outperforms soft attention. \newcite{wu-cotterell-2019-exact} further shows that enforcing monotonicity in a hard attention model improves performance.

\section{Conclusion}
Using a large batch size and feature invariant input allows the transformer to achieve strong performance on character-level tasks. However, it is unclear what linguistic errors the transformer makes compared to recurrent models on these tasks. Future work should analyze the errors in detail as \newcite{gorman-etal-2019-weird} does for recurrent models.
While \citeauthor{wu-cotterell-2019-exact} shows that the monotonicity bias benefits character-level tasks, it is not evident how to enforce monotonicity on multi-headed self-attention.
Future work should consider how to best incorporate monotonicity into the model, either by enforcing it strictly \cite{wu-cotterell-2019-exact} or by pretraining the model to copy \cite{anastasopoulos-neubig-2019-pushing}.

\bibliography{anthology,main}

\begin{thebibliography}{38}
\expandafter\ifx\csname natexlab\endcsname\relax\def\natexlab#1{#1}\fi

\bibitem[{Aharoni and Goldberg(2017)}]{aharoni-goldberg-2017-morphological}
Roee Aharoni and Yoav Goldberg. 2017.
\newblock \href {https://doi.org/10.18653/v1/P17-1183} {Morphological
  inflection generation with hard monotonic attention}.
\newblock In \emph{Proceedings of the 55th Annual Meeting of the Association
  for Computational Linguistics (Volume 1: Long Papers)}, pages 2004--2015,
  Vancouver, Canada. Association for Computational Linguistics.

\bibitem[{Anastasopoulos and Neubig(2019)}]{anastasopoulos-neubig-2019-pushing}
Antonios Anastasopoulos and Graham Neubig. 2019.
\newblock \href {https://doi.org/10.18653/v1/D19-1091} {Pushing the limits of
  low-resource morphological inflection}.
\newblock In \emph{Proceedings of the 2019 Conference on Empirical Methods in
  Natural Language Processing and the 9th International Joint Conference on
  Natural Language Processing (EMNLP-IJCNLP)}, pages 984--996, Hong Kong,
  China. Association for Computational Linguistics.

\bibitem[{Ba et~al.(2016)Ba, Kiros, and Hinton}]{ba2016layer}
Jimmy~Lei Ba, Jamie~Ryan Kiros, and Geoffrey~E. Hinton. 2016.
\newblock Layer normalization.
\newblock \emph{arXiv preprint arXiv:1607.06450}.

\bibitem[{Barrault et~al.(2019)Barrault, Bojar, Costa-juss{\`a}, Federmann,
  Fishel, Graham, Haddow, Huck, Koehn, Malmasi, Monz, M{\"u}ller, Pal, Post,
  and Zampieri}]{barrault-etal-2019-findings}
Lo{\"\i}c Barrault, Ond{\v{r}}ej Bojar, Marta~R. Costa-juss{\`a}, Christian
  Federmann, Mark Fishel, Yvette Graham, Barry Haddow, Matthias Huck, Philipp
  Koehn, Shervin Malmasi, Christof Monz, Mathias M{\"u}ller, Santanu Pal, Matt
  Post, and Marcos Zampieri. 2019.
\newblock \href {https://doi.org/10.18653/v1/W19-5301} {Findings of the 2019
  conference on machine translation ({WMT}19)}.
\newblock In \emph{Proceedings of the Fourth Conference on Machine Translation
  (Volume 2: Shared Task Papers, Day 1)}, pages 1--61, Florence, Italy.
  Association for Computational Linguistics.

\bibitem[{Bergmanis et~al.(2017)Bergmanis, Kann, Sch{\"u}tze, and
  Goldwater}]{bergmanis-etal-2017-training}
Toms Bergmanis, Katharina Kann, Hinrich Sch{\"u}tze, and Sharon Goldwater.
  2017.
\newblock \href {https://doi.org/10.18653/v1/K17-2002} {Training data
  augmentation for low-resource morphological inflection}.
\newblock In \emph{Proceedings of the {C}o{NLL} {SIGMORPHON} 2017 Shared Task:
  Universal Morphological Reinflection}, pages 31--39, Vancouver. Association
  for Computational Linguistics.

\bibitem[{Bollmann(2018)}]{bollmann2018normalization}
Marcel Bollmann. 2018.
\newblock \emph{Normalization of historical texts with neural network models}.
\newblock Ph.D. thesis, Bochum, Ruhr-Universit{\"a}t Bochum.

\bibitem[{Bollmann(2019)}]{bollmann-2019-large}
Marcel Bollmann. 2019.
\newblock \href {https://doi.org/10.18653/v1/N19-1389} {A large-scale
  comparison of historical text normalization systems}.
\newblock In \emph{Proceedings of the 2019 Conference of the North {A}merican
  Chapter of the Association for Computational Linguistics: Human Language
  Technologies, Volume 1 (Long and Short Papers)}, pages 3885--3898,
  Minneapolis, Minnesota. Association for Computational Linguistics.

\bibitem[{Cotterell et~al.(2018)Cotterell, Kirov, Sylak-Glassman, Walther,
  Vylomova, McCarthy, Kann, Mielke, Nicolai, Silfverberg, Yarowsky, Eisner, and
  Hulden}]{cotterell-etal-2018-conll}
Ryan Cotterell, Christo Kirov, John Sylak-Glassman, G{\'e}raldine Walther,
  Ekaterina Vylomova, Arya~D. McCarthy, Katharina Kann, Sebastian Mielke,
  Garrett Nicolai, Miikka Silfverberg, David Yarowsky, Jason Eisner, and Mans
  Hulden. 2018.
\newblock \href {https://doi.org/10.18653/v1/K18-3001} {The
  {C}o{NLL}{--}{SIGMORPHON} 2018 shared task: Universal morphological
  reinflection}.
\newblock In \emph{Proceedings of the {C}o{NLL}{--}{SIGMORPHON} 2018 Shared
  Task: Universal Morphological Reinflection}, pages 1--27, Brussels.
  Association for Computational Linguistics.

\bibitem[{Cotterell et~al.(2017)Cotterell, Kirov, Sylak-Glassman, Walther,
  Vylomova, Xia, Faruqui, K{\"u}bler, Yarowsky, Eisner, and
  Hulden}]{cotterell-etal-2017-conll}
Ryan Cotterell, Christo Kirov, John Sylak-Glassman, G{\'e}raldine Walther,
  Ekaterina Vylomova, Patrick Xia, Manaal Faruqui, Sandra K{\"u}bler, David
  Yarowsky, Jason Eisner, and Mans Hulden. 2017.
\newblock \href {https://doi.org/10.18653/v1/K17-2001} {{C}o{NLL}-{SIGMORPHON}
  2017 shared task: Universal morphological reinflection in 52 languages}.
\newblock In \emph{Proceedings of the {C}o{NLL} {SIGMORPHON} 2017 Shared Task:
  Universal Morphological Reinflection}, pages 1--30, Vancouver. Association
  for Computational Linguistics.

\bibitem[{Cotterell et~al.(2016)Cotterell, Kirov, Sylak-Glassman, Yarowsky,
  Eisner, and Hulden}]{cotterell-etal-2016-sigmorphon}
Ryan Cotterell, Christo Kirov, John Sylak-Glassman, David Yarowsky, Jason
  Eisner, and Mans Hulden. 2016.
\newblock \href {https://doi.org/10.18653/v1/W16-2002} {The {SIGMORPHON} 2016
  shared {T}ask{---}{M}orphological reinflection}.
\newblock In \emph{Proceedings of the 14th {SIGMORPHON} Workshop on
  Computational Research in Phonetics, Phonology, and Morphology}, pages
  10--22, Berlin, Germany. Association for Computational Linguistics.

\bibitem[{Devlin et~al.(2019)Devlin, Chang, Lee, and
  Toutanova}]{devlin-etal-2019-bert}
Jacob Devlin, Ming-Wei Chang, Kenton Lee, and Kristina Toutanova. 2019.
\newblock \href {https://doi.org/10.18653/v1/N19-1423} {{BERT}: Pre-training of
  deep bidirectional transformers for language understanding}.
\newblock In \emph{Proceedings of the 2019 Conference of the North {A}merican
  Chapter of the Association for Computational Linguistics: Human Language
  Technologies, Volume 1 (Long and Short Papers)}, pages 4171--4186,
  Minneapolis, Minnesota. Association for Computational Linguistics.

\bibitem[{Dong et~al.(2019)Dong, Yang, Wang, Wei, Liu, Wang, Gao, Zhou, and
  Hon}]{dong2019unified}
Li~Dong, Nan Yang, Wenhui Wang, Furu Wei, Xiaodong Liu, Yu~Wang, Jianfeng Gao,
  Ming Zhou, and Hsiao-Wuen Hon. 2019.
\newblock Unified language model pre-training for natural language
  understanding and generation.
\newblock In \emph{Advances in Neural Information Processing Systems}, pages
  13042--13054.

\bibitem[{Flachs et~al.(2019)Flachs, Bollmann, and
  S{\o}gaard}]{flachs-etal-2019-historical}
Simon Flachs, Marcel Bollmann, and Anders S{\o}gaard. 2019.
\newblock \href {https://doi.org/10.18653/v1/P19-1157} {Historical text
  normalization with delayed rewards}.
\newblock In \emph{Proceedings of the 57th Annual Meeting of the Association
  for Computational Linguistics}, pages 1614--1619, Florence, Italy.
  Association for Computational Linguistics.

\bibitem[{Gehring et~al.(2017)Gehring, Auli, Grangier, Yarats, and
  Dauphin}]{gehring2017}
Jonas Gehring, Michael Auli, David Grangier, Denis Yarats, and Yann~N. Dauphin.
  2017.
\newblock Convolutional sequence to sequence learning.
\newblock In \emph{Proceedings of the 34th International Conference on Machine
  Learning-Volume 70}, pages 1243--1252. JMLR.

\bibitem[{Gorman et~al.(2019)Gorman, McCarthy, Cotterell, Vylomova,
  Silfverberg, and Markowska}]{gorman-etal-2019-weird}
Kyle Gorman, Arya~D. McCarthy, Ryan Cotterell, Ekaterina Vylomova, Miikka
  Silfverberg, and Magdalena Markowska. 2019.
\newblock \href {https://doi.org/10.18653/v1/K19-1014} {Weird inflects but
  {OK}: Making sense of morphological generation errors}.
\newblock In \emph{Proceedings of the 23rd Conference on Computational Natural
  Language Learning (CoNLL)}, pages 140--151, Hong Kong, China. Association for
  Computational Linguistics.

\bibitem[{Kingma and Ba(2014)}]{kingma2014adam}
Diederik~P. Kingma and Jimmy Ba. 2014.
\newblock Adam: A method for stochastic optimization.
\newblock \emph{arXiv preprint arXiv:1412.6980}.

\bibitem[{Ljube{\v{s}}ic et~al.(2016)Ljube{\v{s}}ic, Zupan, Fi{\v{s}}er, and
  Erjavec}]{ljubevsic2016normalising}
Nikola Ljube{\v{s}}ic, Katja Zupan, Darja Fi{\v{s}}er, and Tomaz Erjavec. 2016.
\newblock Normalising {S}lovene data: historical texts vs. user-generated
  content.
\newblock In \emph{Proceedings of the 13th Conference on Natural Language
  Processing (KONVENS 2016)}, pages 146--155.

\bibitem[{Ljube\v{s}i\'{c} et~al.(2016)Ljube\v{s}i\'{c}, Zupan, Fi{\v s}er, and
  Erjavec}]{ljubesic16-normalising}
Nikola Ljube\v{s}i\'{c}, Katja Zupan, Darja Fi{\v s}er, and Toma{\v z} Erjavec.
  2016.
\newblock {Normalising {S}lovene data: historical texts vs. user-generated
  content}.
\newblock In \emph{Proceedings of the 13th Conference on Natural Language
  Processing (KONVENS 2016)}, pages 146--155.

\bibitem[{Luong et~al.(2015)Luong, Pham, and
  Manning}]{luong-etal-2015-effective}
Thang Luong, Hieu Pham, and Christopher~D. Manning. 2015.
\newblock \href {https://doi.org/10.18653/v1/D15-1166} {Effective approaches to
  attention-based neural machine translation}.
\newblock In \emph{Proceedings of the 2015 Conference on Empirical Methods in
  Natural Language Processing}, pages 1412--1421, Lisbon, Portugal. Association
  for Computational Linguistics.

\bibitem[{Makarov et~al.(2017)Makarov, Ruzsics, and
  Clematide}]{makarov-etal-2017-align}
Peter Makarov, Tatiana Ruzsics, and Simon Clematide. 2017.
\newblock \href {https://doi.org/10.18653/v1/K17-2004} {Align and copy: {UZH}
  at {SIGMORPHON} 2017 shared task for morphological reinflection}.
\newblock In \emph{Proceedings of the {C}o{NLL} {SIGMORPHON} 2017 Shared Task:
  Universal Morphological Reinflection}, pages 49--57, Vancouver. Association
  for Computational Linguistics.

\bibitem[{McCarthy et~al.(2019)McCarthy, Vylomova, Wu, Malaviya, Wolf-Sonkin,
  Nicolai, Kirov, Silfverberg, Mielke, Heinz, Cotterell, and
  Hulden}]{mccarthy-etal-2019-sigmorphon}
Arya~D. McCarthy, Ekaterina Vylomova, Shijie Wu, Chaitanya Malaviya, Lawrence
  Wolf-Sonkin, Garrett Nicolai, Christo Kirov, Miikka Silfverberg, Sebastian~J.
  Mielke, Jeffrey Heinz, Ryan Cotterell, and Mans Hulden. 2019.
\newblock \href {https://doi.org/10.18653/v1/W19-4226} {The {SIGMORPHON} 2019
  shared task: Morphological analysis in context and cross-lingual transfer for
  inflection}.
\newblock In \emph{Proceedings of the 16th Workshop on Computational Research
  in Phonetics, Phonology, and Morphology}, pages 229--244, Florence, Italy.
  Association for Computational Linguistics.

\bibitem[{Pettersson(2016)}]{pettersson2016spelling}
Eva Pettersson. 2016.
\newblock \emph{Spelling normalisation and linguistic analysis of historical
  text for information extraction}.
\newblock Ph.D. thesis, Acta Universitatis Upsaliensis.

\bibitem[{Pettersson et~al.(2013)Pettersson, Megyesi, and
  Tiedemann}]{pettersson2013smt}
Eva Pettersson, Be{\'a}ta Megyesi, and J{\"o}rg Tiedemann. 2013.
\newblock An {SMT} approach to automatic annotation of historical text.
\newblock In \emph{Proceedings of the workshop on computational historical
  linguistics at NODALIDA 2013; May 22-24; 2013; Oslo; Norway. NEALT
  Proceedings Series 18}, 087, pages 54--69. Link{\"o}ping University
  Electronic Press.

\bibitem[{Popel and Bojar(2018)}]{popel2018training}
Martin Popel and Ond{\v{r}}ej Bojar. 2018.
\newblock Training tips for the transformer model.
\newblock \emph{The Prague Bulletin of Mathematical Linguistics},
  110(1):43--70.

\bibitem[{S{\'a}nchez-Mart{\'\i}nez et~al.(2013)S{\'a}nchez-Mart{\'\i}nez,
  Mart{\'\i}nez-Sempere, Ivars-Ribes, and Carrasco}]{sanchez2013open}
Felipe S{\'a}nchez-Mart{\'\i}nez, Isabel Mart{\'\i}nez-Sempere, Xavier
  Ivars-Ribes, and Rafael~C. Carrasco. 2013.
\newblock An open diachronic corpus of historical {S}panish: annotation
  criteria and automatic modernisation of spelling.
\newblock \emph{arXiv preprint arXiv:1306.3692}.

\bibitem[{Scherrer and Erjavec(2013)}]{scherrer2013modernizing}
Yves Scherrer and Toma{\v{z}} Erjavec. 2013.
\newblock Modernizing historical {S}lovene words with character-based smt.
\newblock In \emph{BSNLP 2013-4th Biennial Workshop on Balto-Slavic Natural
  Language Processing}.

\bibitem[{Scherrer and Erjavec(2016)}]{scherrer2016modernising}
Yves Scherrer and Toma{\v{z}} Erjavec. 2016.
\newblock Modernising historical {S}lovene words.
\newblock \emph{Natural Language Engineering}, 22(6):881--905.

\bibitem[{Sejnowski and Rosenberg(1987)}]{Sejnowski1987ParallelNT}
Terrence~J. Sejnowski and Charles~R. Rosenberg. 1987.
\newblock Parallel networks that learn to pronounce {E}nglish text.
\newblock \emph{Complex Systems}, 1.

\bibitem[{Silfverberg et~al.(2017)Silfverberg, Wiemerslage, Liu, and
  Mao}]{silfverberg-etal-2017-data}
Miikka Silfverberg, Adam Wiemerslage, Ling Liu, and Lingshuang~Jack Mao. 2017.
\newblock \href {https://doi.org/10.18653/v1/K17-2010} {Data augmentation for
  morphological reinflection}.
\newblock In \emph{Proceedings of the {C}o{NLL} {SIGMORPHON} 2017 Shared Task:
  Universal Morphological Reinflection}, pages 90--99, Vancouver. Association
  for Computational Linguistics.

\bibitem[{Szegedy et~al.(2016)Szegedy, Vanhoucke, Ioffe, Shlens, and
  Wojna}]{szegedy2016rethinking}
Christian Szegedy, Vincent Vanhoucke, Sergey Ioffe, Jon Shlens, and Zbigniew
  Wojna. 2016.
\newblock Rethinking the inception architecture for computer vision.
\newblock In \emph{Proceedings of the IEEE conference on computer vision and
  pattern recognition}, pages 2818--2826.

\bibitem[{Tang et~al.(2018{\natexlab{a}})Tang, Cap, Pettersson, and
  Nivre}]{tang-etal-2018-evaluation}
Gongbo Tang, Fabienne Cap, Eva Pettersson, and Joakim Nivre.
  2018{\natexlab{a}}.
\newblock \href {https://www.aclweb.org/anthology/C18-1112} {An evaluation of
  neural machine translation models on historical spelling normalization}.
\newblock In \emph{Proceedings of the 27th International Conference on
  Computational Linguistics}, pages 1320--1331, Santa Fe, New Mexico, USA.
  Association for Computational Linguistics.

\bibitem[{Tang et~al.(2018{\natexlab{b}})Tang, M{\"u}ller, Rios, and
  Sennrich}]{tang-etal-2018-self}
Gongbo Tang, Mathias M{\"u}ller, Annette Rios, and Rico Sennrich.
  2018{\natexlab{b}}.
\newblock \href {https://doi.org/10.18653/v1/D18-1458} {Why self-attention? a
  targeted evaluation of neural machine translation architectures}.
\newblock In \emph{Proceedings of the 2018 Conference on Empirical Methods in
  Natural Language Processing}, pages 4263--4272, Brussels, Belgium.
  Association for Computational Linguistics.

\bibitem[{Vaswani et~al.(2018)Vaswani, Bengio, Brevdo, Chollet, Gomez, Gouws,
  Jones, Kaiser, Kalchbrenner, Parmar et~al.}]{vaswani2018tensor2tensor}
Ashish Vaswani, Samy Bengio, Eugene Brevdo, Francois Chollet, Aidan~N Gomez,
  Stephan Gouws, Llion Jones, {\L}ukasz Kaiser, Nal Kalchbrenner, Niki Parmar,
  et~al. 2018.
\newblock Tensor2tensor for neural machine translation.
\newblock \emph{arXiv preprint arXiv:1803.07416}.

\bibitem[{Vaswani et~al.(2017)Vaswani, Shazeer, Parmar, Uszkoreit, Jones,
  Gomez, Kaiser, and Polosukhin}]{vaswani2017attention}
Ashish Vaswani, Noam Shazeer, Niki Parmar, Jakob Uszkoreit, Llion Jones,
  Aidan~N. Gomez, {\L}ukasz Kaiser, and Illia Polosukhin. 2017.
\newblock Attention is all you need.
\newblock In \emph{Advances in neural information processing systems}, pages
  5998--6008.

\bibitem[{Weide(1998)}]{CMUDict}
R.L. Weide. 1998.
\newblock \href {http://www.speech.cs.cmu.edu/cgi-bin/cmudict} {The {C}arnegie
  {M}ellon pronouncing dictionary}.

\bibitem[{Wu and Cotterell(2019)}]{wu-cotterell-2019-exact}
Shijie Wu and Ryan Cotterell. 2019.
\newblock \href {https://doi.org/10.18653/v1/P19-1148} {Exact hard monotonic
  attention for character-level transduction}.
\newblock In \emph{Proceedings of the 57th Annual Meeting of the Association
  for Computational Linguistics}, pages 1530--1537, Florence, Italy.
  Association for Computational Linguistics.

\bibitem[{Wu et~al.(2018)Wu, Shapiro, and Cotterell}]{wu-etal-2018-hard}
Shijie Wu, Pamela Shapiro, and Ryan Cotterell. 2018.
\newblock \href {https://doi.org/10.18653/v1/D18-1473} {Hard non-monotonic
  attention for character-level transduction}.
\newblock In \emph{Proceedings of the 2018 Conference on Empirical Methods in
  Natural Language Processing}, pages 4425--4438, Brussels, Belgium.
  Association for Computational Linguistics.

\bibitem[{Zhang et~al.(2015)Zhang, Li, Banchs, and
  Kumaran}]{zhang-etal-2015-whitepaper}
Min Zhang, Haizhou Li, Rafael~E. Banchs, and A~Kumaran. 2015.
\newblock \href {https://doi.org/10.18653/v1/W15-3901} {Whitepaper of {NEWS}
  2015 shared task on machine transliteration}.
\newblock In \emph{Proceedings of the Fifth Named Entity Workshop}, pages 1--9,
  Beijing, China. Association for Computational Linguistics.

\end{thebibliography}
\bibliographystyle{acl_natbib}

\end{document}